\documentclass[conference]{IEEEtran}
\IEEEoverridecommandlockouts
\usepackage{cite}
\usepackage{amsmath,amssymb,amsfonts}
\usepackage{algorithmic}
\usepackage{graphicx}
\usepackage{textcomp}
\usepackage{xcolor}
\usepackage{lineno,hyperref}

\def\BibTeX{{\rm B\kern-.05em{\sc i\kern-.025em b}\kern-.08em
    T\kern-.1667em\lower.7ex\hbox{E}\kern-.125emX}}
\begin{document}

\title{2DGS-Avatar: Animatable High-fidelity Clothed Avatar via 2D Gaussian Splatting
\thanks{*Corresponding author}
}

\author{\IEEEauthorblockN{Qipeng Yan}
\IEEEauthorblockA{\textit{Academy for Engineering} \\
\textit{and Technology} \\
\textit{Fudan University}\\
Shanghai, China \\
qpyan23@m.fudan.edu.cn}
\and
\IEEEauthorblockN{Mingyang Sun}
\IEEEauthorblockA{\textit{Academy for Engineering} \\
\textit{and Technology} \\
\textit{Fudan University}\\
Shanghai, China \\
mysun21@m.fudan.edu.cn}
\and
\IEEEauthorblockN{Lihua Zhang*}
\IEEEauthorblockA{\textit{Academy for Engineering} \\
\textit{and Technology} \\
\textit{Fudan University}\\
Shanghai, China \\
lihuazhang@fudan.edu.cn}
}

\maketitle

\begin{abstract}
Real-time rendering of high-fidelity and animatable avatars from monocular videos remains a challenging problem in computer vision and graphics.
Over the past few years, the Neural Radiance Field (NeRF) has made significant progress in rendering quality but behaves poorly in run-time performance due to the low efficiency of volumetric rendering. 
Recently, methods based on 3D Gaussian Splatting (3DGS) have shown great potential in fast training and real-time rendering. However, they still suffer from artifacts caused by inaccurate geometry. 
To address these problems, we propose 2DGS-Avatar, a novel approach based on 2D Gaussian Splatting (2DGS) for modeling animatable clothed avatars with high-fidelity and fast training performance.
Given monocular RGB videos as input, our method generates an avatar that can be driven by poses and rendered in real-time.
Compared to 3DGS-based methods, our 2DGS-Avatar retains the advantages of fast training and rendering while also capturing detailed, dynamic, and photo-realistic appearances.
We conduct abundant experiments on popular datasets such as AvatarRex and THuman4.0, demonstrating impressive performance in both qualitative and quantitative metrics.

\end{abstract}


\section{Inroduction}
Creating a high-fidelity and animatable avatar holds significant importance in fields such as AR/VR, entertainment, and film production.
Over the past few years, Neural Radiance Fields (NeRF) \cite{nerf} has been employed by some studies, enabling to reconstruct avatars from videos \cite{peng2021neural, feng2022capturing, jiang2022neuman,su2021anerf} and images \cite{cha2023generating,zhao2022humannerf}. 
Though they achieve photo-realistic rendering, volumetric rendering in NeRF is inefficient and requires expensive training time and computational resources, making them impractical for real-world applications.

Recently, 3D Gaussian Splatting (3DGS) \cite{kerbl20233d} provides a significant solution for fast training and rendering. In contrast to NeRF, 3DGS replaces the hierarchical volume sampling with a depth-based sort along the view direction, named splatting.
Some methods \cite{qian20243dgs,li2024animatable,hu2024gauhuman} adopt 3DGS to model clothed humans, showing great potential in real-time rendering with high visual quality.
However, since 3D Gaussian models use ellipsoids to represent objects, which contradicts the thin nature of surfaces, these approaches may produce fluctuating artifacts and fail to reconstruct accurate geometry.
2D Gaussian Splatting (2DGS) \cite{huang20242d} replaces 3D ellipsoids with 2D ellipses, which are similar to the triangular faces of the mesh, thus it is prone to converge to a more accurate geometry than 3DGS.

Inspired by 2DGS \cite{huang20242d}, we propose 2DGS-Avatar, a novel method for modeling animatable avatars, achieving fast run-time performance and superior geometry quality.
Specifically, the avatar template is first represented by 2D Gaussian primitives in the canonical space, which is initialized by the vertices of the SMPL-X\cite{SMPL-X:2019}. 
Then, the Linear Blend Skinning (LBS) \cite{SMPL-X:2019,SMPL:2015} is employed to transform these 2D Gaussian primitives from the canonical space to the posed space, with each primitive’s skinning weight assigned by querying a diffused skinning weight field \cite{lin2022learning}. 
Finally, we render the RGB images and normal maps with the differentiable rasterizer of 2DGS, which are supervised by the input RGB images and the corresponding normal maps.
For better optimization of 2DGS, a self-supervised loss is put forward to ensure that the Gaussian primitives are uniformly distributed on the surface. 
During the densification phase, we enhance the original strategy in 2DGS with eccentricity filtering that removes the Gaussian primitives with particularly elongated, excessively large, or very low opacity. 

To the best knowledge of us, we are the first work that employs 2DGS to model human avatars.
In summary, our main contributions are as follows:

\begin{itemize}
    \item We introduce 2DGS-Avatar, a novel real-time rendering pipeline for modeling animatable high-fidelity clothed avatars based on 2D Gaussian splatting.
    \item We propose a self-supervised loss that ensures Gaussian primitives are uniformly distributed on the surface, improving rendering results.
    \item We put forward eccentricity filtering to enhance the adaptive density control by removing particularly elongated Gaussian ellipses, resulting in smoother geometric edges.
\end{itemize}

\section{Related Work}
Recently, the emergence of 3DGS \cite{qian20243dgs} has demonstrated impressive capabilities in 3D reconstruction, real-time rendering, and novel view synthesis. This work is also well-suited for representing avatars, leading to various methods\cite{qian20243dgs,li2024animatable,hu2024gauhuman} that extend the 3DGS pipeline to reconstruct human avatars from monocular RGB images. They represent avatars using Gaussian point clouds, optimizing the parameters of the Gaussians for rendering. These approaches can be categorized into two types: learning Gaussian parameters directly and learning Gaussian parameters through 2D maps.
\subsection{Learning Gaussian Parameters Directly}
Methods\cite{qian20243dgs,hu2024gauhuman} that directly learn Gaussian parameters typically follow a pipeline that is similar to NeRF-based approaches\cite{peng2021neural,peng2021animatable,Weng_2022_CVPR}, where the avatar is represented in a canonical space and subsequently transformed into the posed space using LBS, after which the Gaussian primitives are rendered into images through the 3DGS rasterizer. The optimization of Gaussian parameters is performed by minimizing the error between the rendered images and the ground truth, a process that is largely similar to the parameter learning and optimization steps in 3DGS. Additionally, these methods often employ a Multi-Layer Perceptron (MLP) to refine the SMPL pose parameters and LBS skinning weights. Although such methods are characterized by relatively short training times, they are fundamentally limited by the tendency of MLP to prioritize low-frequency information\cite{tancik2020fourfeat}, which consequently hampers their ability to accurately capture high-frequency details such as clothing textures, wrinkles, and other intricate geometric essential for achieving a high level of realism.

\subsection{Learning Gaussian Parameters through 2D Maps}
Methods that learn Gaussian parameters from 2D maps typically representing the human body in the posed space using 2D maps serve as the pose conditions, such as posed position maps\cite{li2024animatable}, UV maps\cite{hu2024gaussianavatar}, and texture maps\cite{Pang_2024_CVPR}. These methods utilize Convolutional Neural Networks (CNN) to directly learn the Gaussian parameters in the posed space. For instance, Animatable Gaussians\cite{li2024animatable} first learns a parametric template, which is then transformed from the canonical space to the posed space using LBS. For each posed space, the template is mapped into front and back posed position maps, and a StyleUNet\cite{wang2023styleavatar} is employed to directly learn and optimize the Gaussian parameters. Subsequently, the 3DGS rasterizer is used to render the images. Due to the use of CNN, these methods are able to capture richer image features, leading to improvements compared to those directly optimizing Gaussian parameters. 
However, because these approaches primarily focus on optimizing CNN, they tend to converge more slowly. Though they can achieve real-time rendering, the training process is more resource-intensive in terms of training time and GPU memory.

\section{Preliminary}
\begin{figure*}[htbp]
\centering
\includegraphics[width=1\textwidth]{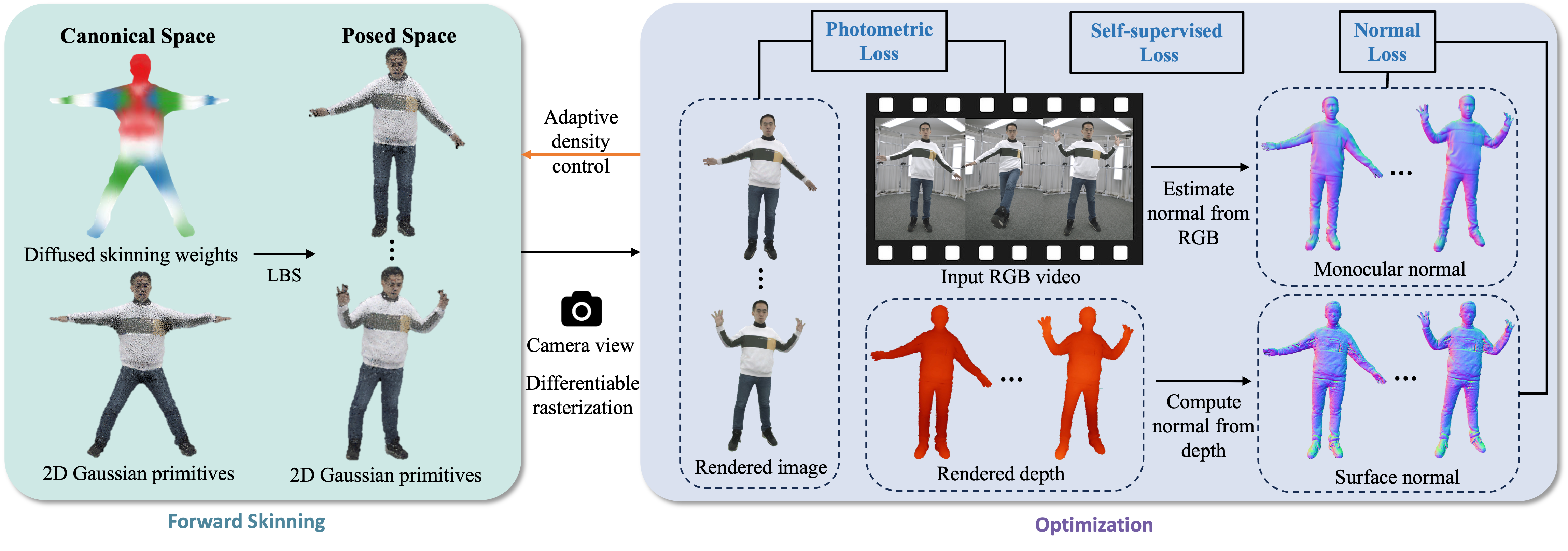}
\caption{Illustration of the pipeline. The orange arrows indicate backpropagation. 
Our method consists of two parts: (1) Transforming Gaussian primitives from the canonical space to the posed space through forward skinning, followed by rasterization to render images and depth maps in the posed space. 
(2) Optimizing the Gaussian primitives in the canonical space using photometric loss, normal loss, and self-supervised loss.
}
\label{pipeline}
\end{figure*}
\subsection{SMPL-X}
SMPL-X \cite{SMPL-X:2019} is a parameterized human model that takes shape parameters $\boldsymbol{\beta}\in\mathbb{R}^{10}$ and pose parameters $\boldsymbol{\theta}\in\mathbb{R}^{K\times 3}$ and returns a triangulated mesh $\mathcal{M}(\beta,\theta)$ by:
\begin{equation}
    \mathcal{M}(\beta,\theta)=\texttt{LBS}(\mathcal W,J(\beta),\theta,T(\beta,\theta)),
\end{equation}
where $\mathcal{M}(\cdot) \in \mathbb{R}^{N \times 3}$, $\texttt{LBS}(\cdot)$ denotes the Linear Blend Skinning (LBS) function, $\mathcal{W}$ is the skinning weights, $J(\cdot)$ represents the joint locations, and $T(\cdot)$ is the template mesh with a star-like rest pose. We set $N = 10475$ and $K=127$, including the body, face, and hands.
In our method, LBS is adopted as a transformation that maps the Gaussian primitives from the canonical space to the posed space.
Specifically, given a point $\mathbf{p}_i$ in the canonical space, the LBS function applies a set of linear transformations to map it to the posed space, resulting in the point $\mathbf{p}_i^\prime$:
\begin{equation}
    \mathbf{p}_i^\prime=\sum_{k=1}^Kw_{k,i}G^\prime_k(\theta,J(\beta))\mathbf{p}_i,
\label{eq1}
\end{equation}
where $w_{k,i}$ is the skinning weight of the $k$-th joint for the $i$-th point, and $G^\prime_k$ is the affine transformation matrix of the $k$-th joint from the canonical space to the posed space.

\subsection{2D Gaussian Splatting}
2DGS \cite{huang20242d} is a novel approach for modeling and reconstructing geometrically accurate radiance fields from multi-view images.
The Gaussian primitives in 3DGS \cite{qian20243dgs} are represented as 3D ellipsoids, while 2DGS flattens the 3D ellipsoid into a 2D ellipse, named surfels.
Each Gaussian primitive is defined by its center point $\mathbf{p}_c\in\mathbb{R}^3$, opacity $\alpha\in\mathbb{R}^1$, 
view-dependent color $\mathbf{c}\in\mathbb{R}^3$ calculated by spherical harmonics coefficients, 
scaling vector $\mathbf{s} = (s_u, s_v)\in\mathbb{R}^2$ that controls the 2D Gaussian variance, and a rotation matrix $\mathbf{R} \in \mathbb{R}^{3 \times 3}$.
The rotation matrix $\mathbf{R} = [\mathbf{r}_u, \mathbf{r}_v, \mathbf{r}_w]$ is composed of two orthogonal vectors $\mathbf{r}_u$ and $\mathbf{r}_v$ on the Gaussian primitive, and the normal vector $\mathbf{r}_w=\mathbf{r}_u\times\mathbf{r}_v$ of the Gaussian primitive, which is obtained by the cross product of these two vectors.
Therefore, a 2D Gaussian is defined in the local tangent plane (also known as the \textit{uv} space) in
world space, which is represented as:
\begin{equation}
    P(u, v)=\mathbf{p}_c+s_u \mathbf{r}_u u+s_v \mathbf{r}_v v=\mathbf{H}(u, v, 1,1)^{\mathrm{T}},
    \label{eq2}
\end{equation}
\begin{equation}
    \mathbf{H}=\left[\begin{array}{cccc}
    s_u \mathbf{r}_u & s_v \mathbf{r}_v & \mathbf{0} & \mathbf{p}_c \\
    0 & 0 & 0 & 1
    \end{array}\right]=\left[\begin{array}{cc}
    \mathbf{R S} & \mathbf{p}_c \\
    \mathbf{0} & 1
    \end{array}\right],
\end{equation}
where $\mathbf{H} \in \mathbb{R}^{4 \times 4}$ represents the geometry of the Gaussian primitive. For a point $\mathbf{u} = (u, v)$ in \textit{uv} space, its Gaussian value $\mathcal{G}(\mathbf{u})$ can be simplified and computed as a Gaussian with a mean of $0$ and a variance of $1$:
\begin{equation}
    \mathcal{G}(\mathbf{u})=\exp\left(-\frac{u^2+v^2}{2}\right).
\end{equation}
Then, its coordinates $P(u, v)$ in the world space can be obtained using \eqref{eq2}.

2DGS maps a pixel $\mathbf{x} = (x, y)$ in screen space to its corresponding point $\mathbf{u} = (u, v)$ in the \textit{uv} space through the Ray-splat Intersection $F$, after which each pixel $\mathbf{c}(\mathbf{x})$ is computed using alpha blending:
\begin{equation}
    \mathbf{c}(\mathbf{x})=\sum_{i=1} \mathbf{c}_i \alpha_i \mathcal{G}_i(F(\mathbf{x})) \prod_{j=1}^{i-1}\left(1-\alpha_j \mathcal{G}_j(F(\mathbf{x}))\right),
\end{equation}
where $\mathbf{c}_i$ represents the color of the Gaussian primitive calculated by spherical harmonic. In summary, the learnable parameters of $\mathcal{G}_i$ are $\Theta_i = \{\mathbf{p}_i, \mathbf{s}_i, \mathbf{R}_i,\alpha_i, \mathbf{c}_i\}$.

\section{Method}
Given multi-view RGB videos and the related SMPL-X registrations that include the pose and shape parameters for the character in each frame, our goal is to create an animatable high-fidelity clothed avatar. 
The pipeline is shown in Fig. \ref{pipeline}.

First, we precompute a skinning weight field\cite{lin2022learning} in the canonical space, diffusing the skinning weights from the SMPL-X surface to the entire canonical space.
This allows us to obtain the skinning weight of each Gaussian primitive by querying the skinning weight field.
Second, we transform the Gaussian primitives from the canonical space to the posed space using \eqref{eq1}, followed by rasterization to render images and depth maps of the Gaussian primitives in the posed space.
We optimize the model by minimizing the photometric difference between the rendered images and the corresponding frames of the input RGB videos, as well as the difference between the normals computed from the depth maps and those estimated from the input RGB images.
Additionally, we propose a self-supervised loss to constrain the distribution of Gaussian primitives and the smoothness of the normal maps. Finally, we propose an eccentricity filtering algorithm to control the density adaptively.

\subsection{Foward Skinning}
We initialize a set of Gaussian primitives $\{\mathcal{G}_i\}_{i=1}^N$ at the vertices of the SMPL-X model in the canonical space according to the shape parameters $\boldsymbol{\beta}$ from the dataset.
We then query the precomputed diffused skinning weight field using the center points $\mathbf{p}_c$ of the Gaussian primitives $\mathcal{G}_i$ to obtain the corresponding LBS weights $w_i$.
After each optimization step, the skinning weights are re-queried.
The center point $\mathbf{p}_c$ is transformed from the canonical space to the corresponding point $\mathbf{p}_c'$ in the posed space by  \eqref{eq1}.

\subsection{Splatting}
Based on the camera’s intrinsic and extrinsic parameters from the input RGB images, the rendered image can be obtained using alpha blending $\mathbf{c}(\mathbf{x}) = \sum_{i=1}^N \mathbf{c}_i \alpha_i \mathcal{G}_i(F(\mathbf{x}))T_i$, where $T_i = \prod_{j=1}^{i-1} \left(1 - \alpha_j \mathcal{G}_j(F(\mathbf{x}))\right)$ represents the accumulated transmittance along the ray from $\mathcal{G}_1$ to $\mathcal{G}_{i-1}$, indicating the visibility of $\mathcal{G}_i$.
Similar to 2DGS, we consider $T_i = 0.5$ as the surface.
Therefore, we only take the depth maps of the visible surface, defined as $z = \max\{z_i \mid T_i > 0.5\}$.
Based on the depth map, we can also compute the normal vectors.
Specifically, for a point $\mathbf{p}$ in the depth map, with its neighboring points along the x-axis $\mathbf{p}_x$ and the y-axis $\mathbf{p}_y$, the normal vector $\mathbf{n}$ can be calculated using the following equation:
\begin{equation}
    \mathbf{n}=\texttt{Normalize}((\mathbf{p}-\mathbf{p}_x)\times(\mathbf{p}-\mathbf{p}_y)).
\end{equation}

\subsection{Optimization}
To optimize the parameters $\Theta$ of the Gaussian primitives $\mathcal{G}$, our loss function $\mathcal{L}$ consists of four components: the photometric loss $\mathcal{L}_p$, the normal loss $\mathcal{L}_n$, the self-supervised loss $\mathcal{L}_s$, and the mask loss $\mathcal{L}_m$. The total loss $\mathcal{L}$ is given by:
\begin{equation}
    \mathcal{L}=\mathcal{L}_p+\lambda_n\mathcal{L}_n+\lambda_s\mathcal{L}_s+\lambda_m\mathcal{L}_m.
\end{equation}

\textbf{Photometric Loss.} The photometric loss consists of two terms same to 2DGS: an $L_1$ term and a D-SSIM term.
In addition to these, we include an additional term, the Learned Perceptual Image Patch Similarity (LPIPS)\cite{zhang2018unreasonable}, to minimize the difference between the rendered image $\hat{\mathbf{I}}$ and the input image $\mathbf{I}$:
\begin{equation}
    \mathcal{L}_p=L_1(\hat{\mathbf{I}},\mathbf{I})+\lambda_{dssim}L_{dssim}(\hat{\mathbf{I}},\mathbf{I})+\lambda_{lpips}L_{lpips}(\hat{\mathbf{I}},\mathbf{I}).
\end{equation}

\textbf{Normal Loss.} Using only the photometric loss is insufficient for accurately modeling human geometry, especially in high-frequency regions.
Therefore, we use the normal loss as a prior.
We apply PIFuHD\cite{saito2020pifuhd} to infer the normal map $\mathbf{N}$ from the input RGB image and compute the normal map $\hat{\mathbf{N}}$ from the rendered depth map.
The geometry of the avatar is constrained by minimizing the cosine similarity between two normal maps:
\begin{equation}
    \mathcal{L}_n=\lambda_n(1-\hat{\mathbf{N}}\cdot\mathbf{N}).
\end{equation}

\begin{figure*}[tbp]
\centering
\includegraphics[width=1\linewidth]{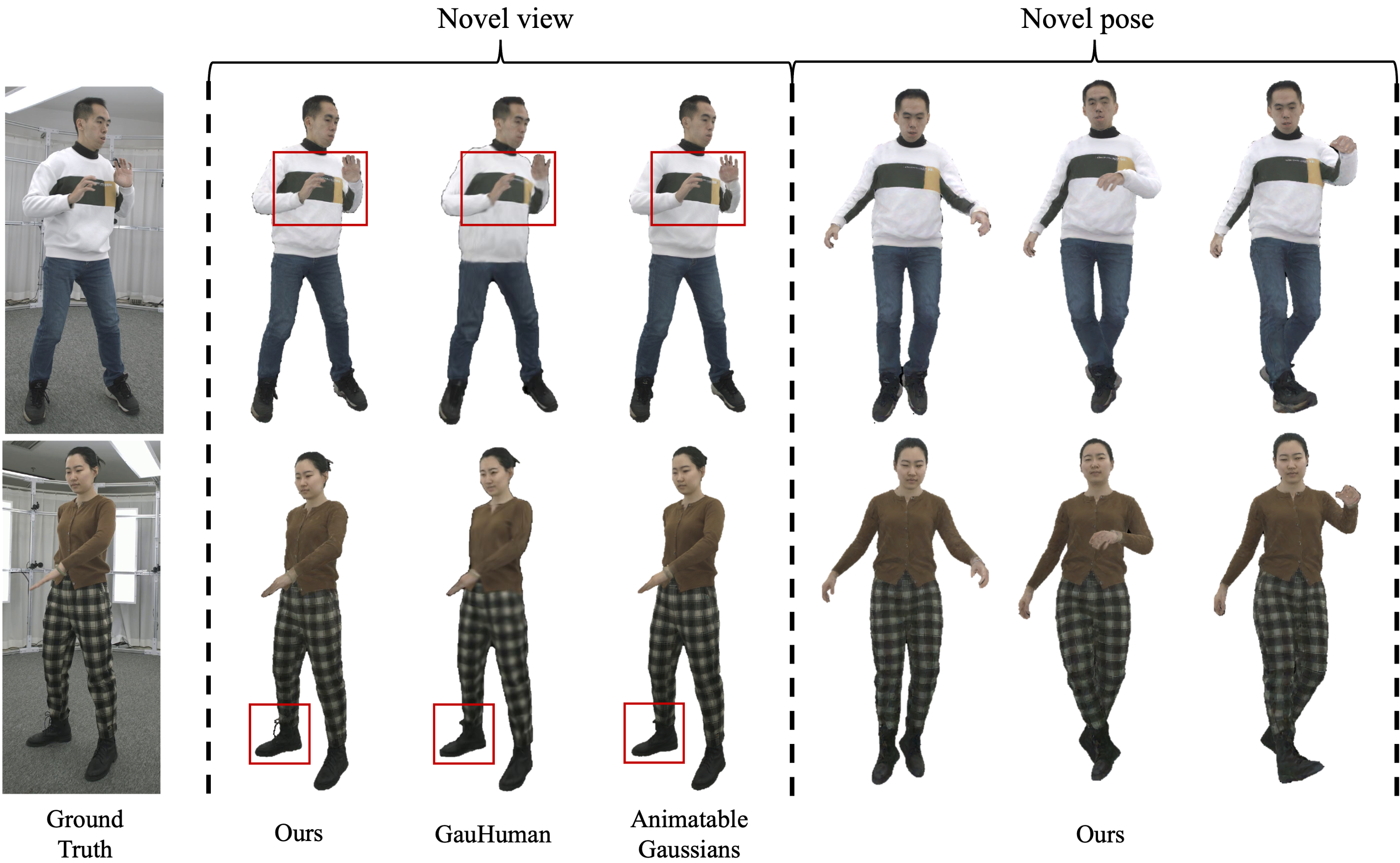}
\caption{Qualitative comparison on AvatarRex\cite{zheng2023avatarrex}. We show the results for both novel view and novel pose on sequences of ``avatarrex\_zzr'' and ``avatarrex\_lbn2'' in AvatarRex. Our method reaches comparable visual effects to Animatable Gaussians \cite{hu2024gaussianavatar} while surpassing GauHuman \cite{hu2024gauhuman} in terms of surface details, such as hands, clothes and shoes.
}
\label{example2}
\end{figure*}

\textbf{Self-supervised Loss.}
Gaussian primitives are typically not uniformly distributed, there are more Gaussian primitives in high-frequency regions and fewer in low-frequency regions.
Since our LBS weights are diffused from the SMPL model, which is inherently designed for meshes, we propose a self-supervised area loss $L_{area}$ to constrain the scaling of each Gaussian primitive by minimizing the variance of the product of the two scaling vectors $(s_u, s_v)$.
This approach ensures that the Gaussian primitives are distributed evenly, similar to triangular faces of mesh.
Additionally, we observed that there are some Gaussian primitives with low opacity inside or outside the avatar, which are not meaningful.
To address this, we introduce an opacity loss $L_{opacity}$ inspired by Gaussian surfels\cite{dai2024high}, encouraging the opacity of the Gaussian primitives to be either close to 1 or 0, thereby ensuring that all Gaussian primitives are distributed on the surface of the avatar:
\begin{equation}
    \mathcal{L}_s=\lambda_{area}L_{area}+\lambda_{opacity}L_{opacity},
\end{equation}
where $L_{opacity}=\exp{\left(-(\alpha_i-0.5)^2/0.05\right)}$.

\textbf{Mask Loss.}
Following NeuS\cite{wang2021neus}, we include a mask loss $\mathcal{L}_m$, which is obtained by calculating the binary cross-entropy between the alpha map $\hat{\mathbf{M}}=\sum_{i=1}^N \alpha_i \mathcal{G}_i(F(\mathbf{x}))T_i$ and the mask $\mathbf{M}$ from the dataset:
\begin{equation}
    \mathcal{L}_m=\lambda_m\texttt{BCE}(\hat{\mathbf{M}},\mathbf{M}).
\end{equation}

\textbf{Eccentricity Filtering.} The previous area loss only constrains the area of the Gaussian primitives, but this can still result in some very elongated ellipses, which may lead to unsmooth geometry at the edges.
Therefore, we propose eccentricity filtering for adaptive density control, which removes Gaussian primitives with an eccentricity higher than a threshold that we set to 9.

\section{Experiments}
\begin{figure}[tbp]
\centering
\includegraphics[width=1\linewidth]{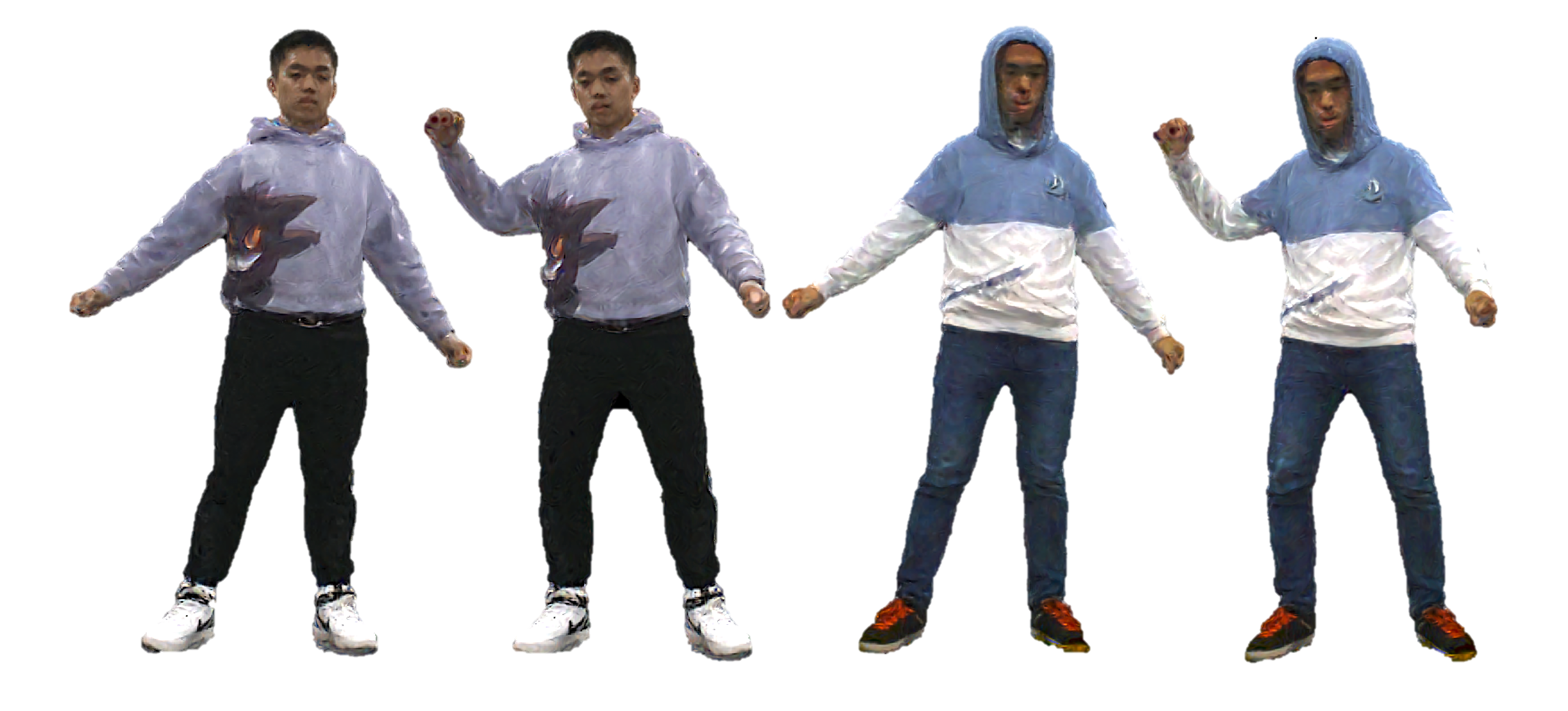}
\caption{More results on sequences of ``subject00'' and ``subject02'' in THuman4.0\cite{zheng2022structured} with novel pose.}
\label{example}
\end{figure}

\subsection{Experimental Setup}
\textbf{Datasets.} Our experiments are conducted on two popular datasets, AvatarRex\cite{zheng2023avatarrex} and THuman4.0\cite{zheng2022structured}, including 4 sequences with 16 views from the AvatarRex and 3 sequences with 24 views from the THuman4.0.
From each dataset, we select 2 sequences for evaluation.
Both datasets provide RGB images, masks, and SMPL-X registrations.
In addition, we utilize PIFuHD \cite{saito2020pifuhd} to estimate normal maps from the RGB images for further supervision in our pipeline.

\textbf{Metric.} We select Peak Signal-to-Noise Ratio (PSNR), Structural Similarity Index Measure (SSIM)\cite{1284395}, Learned Perceptual Image Patch Similarity (LPIPS)\cite{zhang2018unreasonable} as well as training time on an NVIDIA V100 GPU as quantitative metrics for comparative experiments.

\textbf{Baselines.} We compare our approach with state-of-the-art 3DGS-based methods\cite{hu2024gauhuman,li2024animatable} from two categories: learning Gaussian parameters directly and learning Gaussian parameters through 2D Maps.
For each sequence, we split the data into training and testing sets.
All methods are run for 30,000 iterations under their respective default parameter settings.
We report the average metric values for the selected sequences across all methods.

\begin{table}[tbp]
    \centering
    \caption{Quantitative comparison with state-of-the-art methods.}
    \begin{tabular}{llllc}
        \hline Method & PSNR $\uparrow$ & SSIM $\uparrow$ & LPIPS $\downarrow$ & Train \\
        \hline Ours & \underline{30.93} & \underline{0.9643} & \underline{31.37} & \textbf{1 h} \\
        GauHuman\cite{hu2024gauhuman} & 29.57 & 0.9639 & 35.93 & \textbf{1 h} \\
        Animatable Gaussians\cite{li2024animatable} & \textbf{31.86} & \textbf{0.9705} & \textbf{29.32} & 7 h \\
        \hline
        \multicolumn{5}{l}{$\bullet$ The best results are shown in \textbf{bold}, while the second best performance} \\
        \multicolumn{5}{l}{is \underline{underlined}.}
    \end{tabular}
    \label{tab1}
\end{table}

\subsection{Results}
\textbf{Reconstruction.} Fig. \ref{example2} and Table \ref{tab1} summarize the comparison results between our 2DGS-Avatar, GauHuman \cite{hu2024gauhuman}, and Animatable Gaussians \cite{hu2024gaussianavatar}.
Since GauHuman preloads the images into memory before training, a process that takes approximately 30 minutes, rather than using a generator like other methods, we included this time as part of its training time for fairness.
As shown in Fig. \ref{example2}, both our method and Animatable Gaussians produce visibly superior rendering quality compared to GauHuman, with clearer textures in clothing and facial features.
This is because GauHuman relies solely on MLP to learn image features, which limits its ability to optimize Gaussian properties effectively.
From Table \ref{tab1}, it can be observed that both our method and GauHuman require only one hour of training time.
However, our approach and Animatable Gaussians outperform GauHuman in all quantitative metrics.
While our method falls slightly behind Animatable Gaussians, we achieve similar results in only one-seventh of the training time, with minimal perceptible differences in the rendered images.
Additionally, Animatable Gaussians require significantly more GPU memory compared to our approach, further highlighting the efficiency of our method.

\textbf{Animation.} As shown in Fig. \ref{example2} and Fig. \ref{example}, the reconstructed avatar can be driven by LBS with novel pose sequences sampled from AMASS \cite{mahmood2019amass} and  THuman4.0\_POSE\cite{zheng2022structured}. The run-time performance can reach 60 FPS on a single NVIDIA RTX-4070s GPU, which is competent in real-world applications.

\subsection{Ablation Study}
\begin{figure}[tbp]
\centering
\includegraphics[width=1\linewidth]{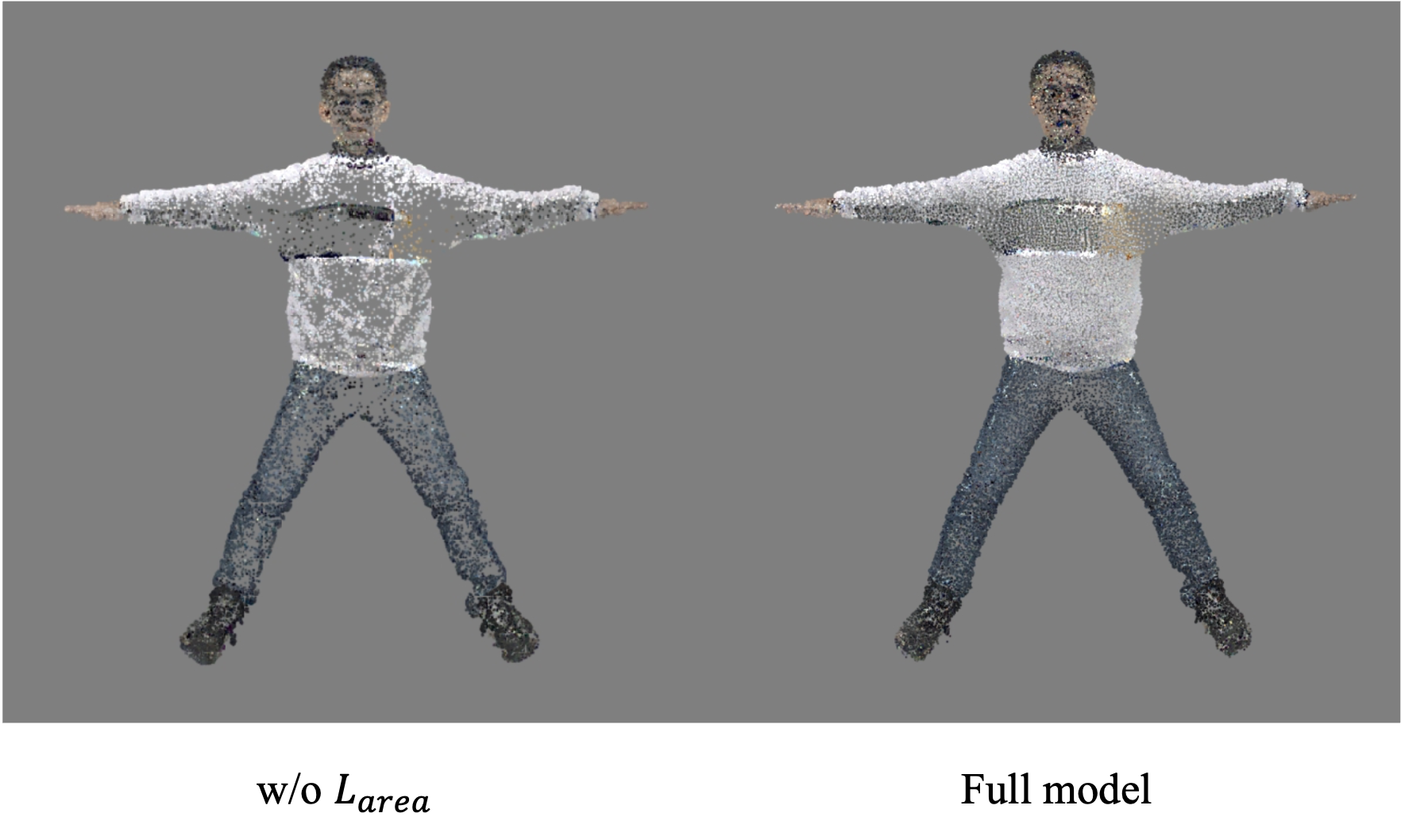}
\caption{The visualization of the ablation study on $L_{area}$. With $L_{area}$, the Gaussian primitives are prone to converge towards a more uniform distribution around the surface.}
\label{ablation}
\end{figure}

We study the effect of different components proposed in our method, including the area loss, normal loss, and eccentricity filtering strategy.
The metrics are presented in the Table \ref{tab2}.
The full model achieves the best results across all quantitative metrics, demonstrating that all the proposed modules are effective, and optimal performance is only achieved when all modules work together. 
In addition, we also visualize the effect of $\mathcal{L}_{area}$ in Fig. \ref{ablation}. The results show that our proposed $\mathcal{L}_{area}$ effectively leads to a more uniform distribution of Gaussian primitives.
\begin{table}[tbp]
    \centering
    \caption{Ablation study on AvatarRex\cite{zheng2023avatarrex}.}
    \begin{tabular}{lccc}
        \hline  & PSNR $\uparrow$ & SSIM $\uparrow$ & LPIPS $\downarrow$ \\
        \hline Full model & $\textbf{31.36}$ & $\textbf{0.9781}$ & $\textbf{34.75}$ \\
        w/o $L_{area}$  & 31.35  & 0.9775 & 36.56 \\
        w/o $L_{normal}$ & 31.26 & 0.9773 & 35.22 \\
        w/o eccentricity filtering & 31.19 & 0.9772 & 35.83 \\
        \hline
        \multicolumn{4}{l}{$\bullet$ The best results are shown in \textbf{bold}.}
    \end{tabular}
    
    \label{tab2}
\end{table}

\section{Discussion}
\textbf{Conclusion.} In this paper, we introduced 2DGS-Avatar, which, to the best of our knowledge, is the first method to represent clothed avatars using 2DGS.
Our approach efficiently reconstructs high-fidelity clothed avatars from monocular RGB videos and enables real-time rendering.
Experimental results demonstrate that our method strikes an effective trade-off between rendering quality, memory consumption and training time, achieving near state-of-the-art performance with significantly reduced computational resources and time.

\textbf{Limitations.} 
(1) When the input RGB videos are relatively blurry, the reconstruction quality tends to degrade, particularly in shadowed areas where lighting is insufficient. Gaussian-DK \cite{ye2024light} alleviates this problem by extracting a lightness map.
(2) Although 2DGS-Avatar is capable of reconstructing high-fidelity avatars, the animation of the avatar, particularly in simulating clothing wrinkles, lacks a high level of realism. IF-Garments \cite{sun2024if} shows great potential in modeling clothing details by combining neural fields with XPBD \cite{macklin2016xpbd}.
(3) It still remains a challenge to model the less frequently observed regions, such as underarms or shoe soles.

\bibliographystyle{IEEEtran}
\bibliography{reference}

\end{document}